%% file: neurips_2025.tex
\documentclass{article}

\usepackage[preprint]{neurips_2025}

\usepackage[utf8]{inputenc} 
\usepackage[T1]{fontenc}    
\usepackage{hyperref}       
\usepackage{url}            
\usepackage{booktabs}       
\usepackage{amsfonts}       
\usepackage{amsmath}        
\usepackage{nicefrac}       
\usepackage{microtype}      
\usepackage{xcolor}         
\usepackage{graphicx}

\usepackage{algorithm}
\usepackage{algpseudocode}

\usepackage{amssymb}
\usepackage{wrapfig}
\usepackage{booktabs}
\usepackage{graphicx}
\usepackage{caption}
\usepackage{geometry}
\usepackage{array}
\usepackage{multirow}
\usepackage{makecell} 
\usepackage{subfig}

\definecolor{DeepGreen}{rgb}{0.0, 0.75, 0.0}

\newcommand{\methodshort}[1]{\textsc{FlashVLA}}

\title{Think Twice, Act Once: Token-Aware Compression and Action Reuse for Efficient Inference in Vision-Language-Action Models}

\author{
    \textbf{Xudong Tan}$^1$$^\dagger$,
    \textbf{Yaoxin Yang}$^1$$^\dagger$,
    \textbf{Peng Ye}$^{2,3}$\thanks{Corresponding Authors: 20110720039@fudan.edu.cn, eetchen@fudan.edu.cn.~~~$^\dagger$Equal Contribution.},
    \textbf{Jialin Zheng}$^1$,\\
    \textbf{Bizhe Bai}$^1$,
    \textbf{Xinyi Wang}$^1$,
    \textbf{Jia Hao}$^4$,
    \textbf{Tao Chen}$^1$\footnotemark[1]
    \\
    $^1$ Fudan University \quad $^2$ Shanghai AI Laboratory \\ $^3$ The Chinese University of Hong Kong \quad $^4$ Zhangjiang Laboratory
}

\begin{document}

\maketitle

\input{sec/0_Abstract}

\input{sec/1_Introduction}

\input{sec/2_Related_Work}

\input{sec/3_Method}
\input{sec/4_Experiments}

\input{sec/5_Conclusion}

{
\small
\bibliographystyle{abbrv}
\bibliography{main}
}

\newpage
\appendix
\input{sec/appendix}

\end{document}

%% file: sec/0_Abstract.tex
\begin{abstract}
Vision-Language-Action (VLA) models have emerged as a powerful paradigm for general-purpose robot control through natural language instructions. However, their high inference cost—stemming from large-scale token computation and autoregressive decoding—poses significant challenges for real-time deployment and edge applications. While prior work has primarily focused on efficient architectural optimization, we take a different and innovative perspective by identifying a dual form of redundancy in VLA models: (i) high similarity across consecutive action steps, and (ii) substantial redundancy in visual tokens. Motivated by these observations, we propose \methodshort{}, the first training-free and plug-and-play acceleration framework that enables action reuse in VLA models. Specifically, \methodshort{} improves inference efficiency through a token-aware action reuse mechanism that avoids redundant decoding across stable action steps, and an information-guided visual token selection strategy that prunes low-contribution tokens. Extensive experiments on the LIBERO benchmark show that \methodshort{} reduces FLOPs by 55.7\% and latency by 36.0\%, with only a 0.7\% drop in task success rate. These results demonstrate the effectiveness of \methodshort{} in enabling lightweight, low-latency VLA inference without retraining.
\end{abstract}

%% file: sec/1_Introduction.tex
\section{Introduction}
\label{sec:Introduction}
In the development of embodied intelligence systems, Vision-Language-Action (VLA) models are rapidly emerging as a key technology for enabling general-purpose behavior control. By integrating visual perception, language understanding, and action generation, VLA models empower embodied agents to execute complex tasks based on natural language instructions, demonstrating strong generalization and task adaptability~\cite{brohan2022rt,nair2022r3m,bai2023qwen,chen2023executing,cheang2024gr,li2023vision,jiang2023motiongpt,kim2024openvla,chen2024ll3da,chi2023diffusion,duan2024manipulate,singh2023progprompt}. However, despite their impressive performance in task execution, VLA models often involve heavy computational loads and high latency during inference, making them difficult to deploy in high-frequency control settings and limiting their applicability in more dexterous and complex bimanual manipulation tasks~\cite{kim2025openvlaoft,liu2024rdt,wen2025tinyvla,li2024inference}.

Current VLA architectures typically fall into two paradigms: the autoregressive generation paradigm, such as OpenVLA~\cite{kim2024openvla}, which encodes multimodal inputs into tokens and decodes actions step-by-step using a language model; and the diffusion policy paradigm~\cite{wen2024diffusion,li2024cogact,yan2024dnact,chi2023diffusion,hou2024diffusion}, which formulates action generation as a conditional denoising process and enables parallel trajectory sampling. Both paradigms rely heavily on Transformer architectures, where each inference step incurs a quadratic complexity $\mathcal{O}(N^2)$ with respect to sequence length $N$, leading to high computational cost. To mitigate this, recent work focuses on architectural-level optimizations, including action chunking~\cite{song2025pdvla,liu2024bidirectional}, parallel decoding~\cite{kim2025openvlaoft}, low-rank adaptation~\cite{wen2025tinyvla}, and model quantization~\cite{park2024quantization}. While effective, these methods typically require additional training overhead.

\begin{figure*}
    \centering
    \includegraphics[width=0.95\textwidth]{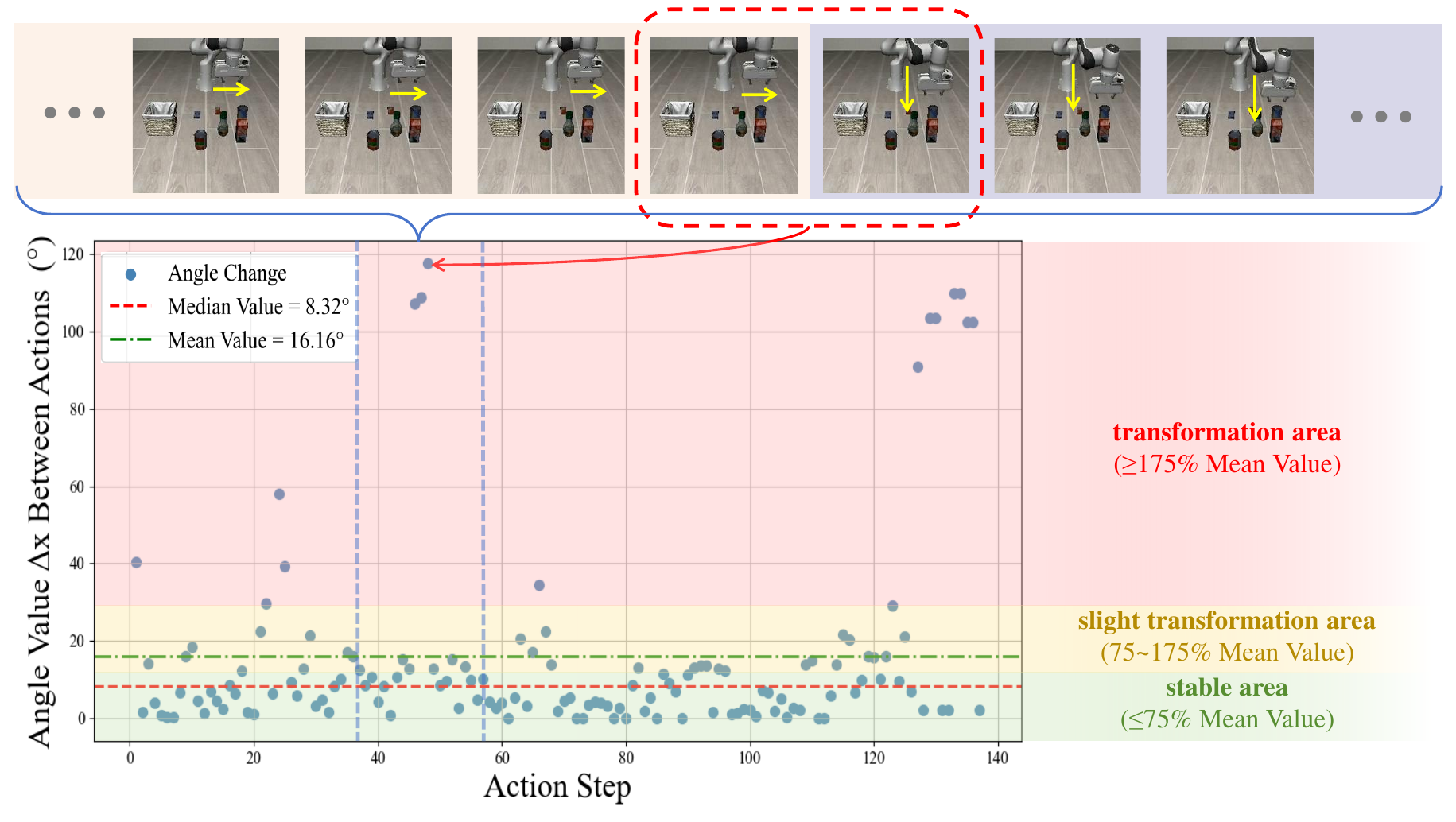}
    \vspace{-0.2cm}
    \caption{Motivation behind our proposed \methodshort{}. The figure shows the change in the VLA model’s output vector at each time step relative to the previous one. The vertical axis indicates the directional difference between consecutive actions. Most actions remain highly consistent with the previous step and appear in the stable area of the figure, while only a few exhibit significant changes.}
    \label{fig1_Motivation}

\end{figure*}

Unlike prior work that focuses on architectural optimizations, we take a new perspective by analyzing the temporal behavior of VLA models at the action level. We observe that in many tasks (e.g., OpenVLA~\cite{kim2024openvla}), \textbf{consecutive action steps often show minimal directional change, suggesting semantic redundancy} (see Fig.~\ref{fig1_Motivation}). This indicates that actions in stable phases can be reused to avoid redundant computation. We further find \textbf{significant redundancy in visual tokens} (see Appendix~\ref{sec:app_Visual_Redundancy}), consistent with observations in vision-language models (VLMs)~\cite{chen2024image,yang2024visionzip,zhang2025llava,chen2024efficient}, which reveals potential for reducing computational cost along another computational dimension.

Building on these observations, we propose \methodshort{}, a training-free, plug-and-play dual-path acceleration framework that improves VLA inference efficiency via action reuse and token pruning. The first component is a \textbf{token-aware reuse mechanism} that compares both action similarity and visual token stability to decide whether to skip computation and reuse the previous action. The second is a \textbf{visual token selection strategy} based on information contribution scores, retaining informative tokens while discarding low-impact ones. The proposed \methodshort{} integrates seamlessly with Flash Attention-based VLA models and follows a ``\textbf{Think Twice, Act Once}'' paradigm: it performs lightweight assessment before execution, selecting between skipped execution and lightweight execution, significantly reducing FLOPs and latency while maintaining control accuracy.

To evaluate the effectiveness of \methodshort{}, we conduct systematic experiments on four representative tasks from the LIBERO benchmark: Spatial, Object, Goal, and Long. We further validate the method on VLAbench to assess generalization. \methodshort{} achieves a 55.7\% reduction in FLOPs and a 36.0\% reduction in latency without any additional training, while reducing the number of visual tokens to 62.5\% of the original input. Notably, the average success rate drops by only 0.7\% compared to the baseline VLA model. Ablation studies and benchmark results collectively demonstrate that \methodshort{} significantly reduces computational cost while preserving task performance, enabling efficient VLA inference with minimal performance drop.

Our key contributions are summarized as follows:
\begin{enumerate}
\item We identify a novel form of action-level and token-level redundancy in VLA inference. Specifically, we observe that most consecutive action steps yield highly similar outputs with only minor directional changes, allowing action reuse in stable phases. Additionally, many visual tokens contribute little to the overall inference process, revealing a degree of visual redundancy similar to that observed in MLLMs.

\item We introduce \methodshort{}, the first training-free and plug-and-play acceleration framework that enables action reuse in VLA models. It integrates a token-aware action reuse mechanism to skip redundant computation in stable action steps, and a visual token selection strategy based on information contribution scores to retain informative tokens. It is worth noted that \methodshort{} is fully compatible with Flash Attention-based VLA backbones.

\item We conduct comprehensive experiments on four representative tasks from the LIBERO benchmark. When visual tokens are reduced to 62.5\% of the original input, \methodshort{} lowers inference latency by 36.0\% and decreases the FLOPs of visual token computation by 55.7\%, while incurring only a 0.7\% drop in success rate. These results demonstrate that \methodshort{} achieves significant efficiency gains with limited performance trade-off.

\end{enumerate}

%% file: sec/2_Related_Work.tex
\section{Related Work}
\label{sec:Related_Work}

Recent advances in VLA models highlight the critical role of architecture in determining both performance and efficiency. To address the computational challenges inherent in these models, a variety of acceleration methods~\cite{kim2025openvlaoft,song2025pdvla,li2024cogact,wen2025tinyvla,liu2024robomamba,xu2025vlacache,yue2024deer,park2024quantization} have been proposed across two dominant paradigms: autoregressive generation and diffusion policy~\cite{zitkovich2023rt,kim2024openvla,black2410pi0,liu2024rdt,wen2024diffusion,yan2024dnact,li2024cogact,chen2023executing}.

\paragraph{Vision-language-action models.}
Vision-language-action (VLA) models provide a promising direction for training generalist robot policies~\cite{ahn2022can,brohan2022rt,black2410pi0,duan2024manipulate} and are built on large-scale robot learning datasets~\cite{liu2023libero,o2024open,fang2023rh20t,khazatsky2024droid,li2024evaluating}. Most VLA models follow one of two paradigms: autoregressive generation and diffusion policy. Autoregressive models, such as RT-2~\cite{zitkovich2023rt} and OpenVLA~\cite{kim2024openvla}, encode multimodal inputs into tokens and decode actions step-by-step using a language model. RT-2 treats actions as text tokens and trains them alongside natural language, while OpenVLA combines a vision backbone with a language model trained on large-scale robot trajectories. Pi0~\cite{black2410pi0} is another autoregressive model that uses flow matching for faster action generation. Diffusion-based models, including RDT-1B~\cite{liu2024rdt}, Diffusion-VLA~\cite{wen2024diffusion}, DNACT~\cite{yan2024dnact}, and CogACT~\cite{li2024cogact}, formulate action generation as conditional denoising. Diffusion-VLA combines autoregressive and diffusion methods to improve robustness. DNACT focuses on multi-task 3D policy learning, while CogACT generates diverse action sequences to improve flexibility. However, the large size of VLA models leads to high computational cost, limiting real-time deployment and high-frequency control.

\paragraph{Acceleration for VLA models.}
Existing methods mainly focus on architectural-level optimizations tailored to the two main VLA paradigms, including action chunking~\cite{song2025pdvla,kim2025openvlaoft}, which splits complex actions into smaller segments to reduce per-step computation; parallel decoding~\cite{song2025pdvla,kim2025openvlaoft}, which enables simultaneous generation of multiple actions; low-rank adaptation~\cite{wen2025tinyvla,hu2022lora}, which compresses model weights to reduce parameters; and model quantization~\cite{pertsch2025fast,park2024quantization}, which lowers numerical precision to save memory and computation. While these techniques improve efficiency, they require additional training or fine-tuning.
Training-free acceleration remains underexplored. Although some pruning methods from VLMs~\cite{chen2024fastv,zhang2024sparsevlm,yang2024visionzip,liu2024multi} are training-free, they are incompatible with Flash Attention and do not sufficiently consider the structural characteristics of VLA models.
To address these limitations, we propose a training-free, plug-and-play dual-path framework that accelerates VLA inference through action reuse and token pruning.

%% file: sec/3_Method.tex
\definecolor{myblue}{RGB}{0,176,240}  
\definecolor{mypurple}{RGB}{112,48,160}  

\begin{figure*}
    \centering
    \includegraphics[width=\linewidth]{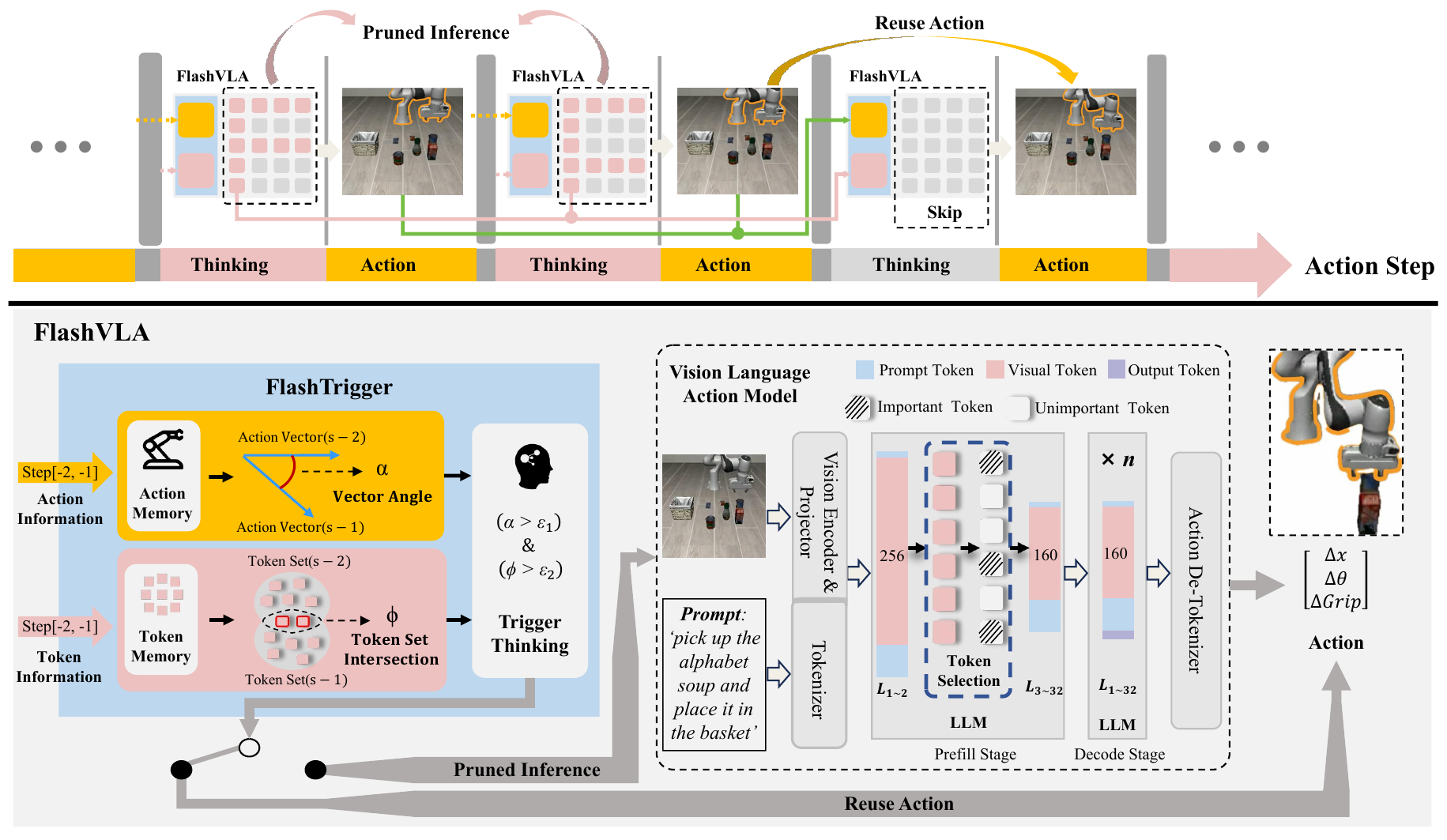}
    \vspace{-0.2cm}
    \caption{Framework of our \methodshort{}. We give the way our method works as the action step changes. Before each inference, FlashTrigger will think about whether it can reuse the output of the previous action based on action memory and token memory (as shown in blue block). If the trigger condition is met, this inference is skipped. If the trigger condition is not met, proceed to the pruned inference step. In pruned inference step, we select the set of important visual tokens in the prefill stage and prune the other unimportant tokens. After inference, action information and token information are used to update action memory and token memory.}
    \label{fig2_Framework}
    \vspace{-0.1cm}
\end{figure*}


%

\section{\methodshort{}}

\label{sec:Method}
The \methodshort{} architecture is illustrated in Fig.\ref{fig2_Framework}. In the following, we first introduce the standard formulation of VLA models in Section\ref{Preliminaries}. We then detail our visual token selection strategy and action reuse mechanism in Section~\ref{Visual Token Selection Strategy via Information Contribution Theory} and Section~\ref{Token-Aware Action reuse strategy}, respectively. The overall algorithmic flow of \methodshort{} is provided in Appendix~\ref{sec:app_algo} (Algorithm~\ref{alg}).

\subsection{Preliminaries}

\label{Preliminaries}

VLA models extend vision-and-language foundation models for robotic control by generating actions conditioned on visual inputs and language prompts. A representative example is OpenVLA~\cite{kim2024openvla}, a 7B-parameter open-source model that establishes a strong baseline for general-purpose manipulation. It consists of a vision encoder that combines DINOv2~\cite{oquab2023dinov2} and SigLIP~\cite{zhai2023sigmoid} features, a projector that maps visual features into the language embedding space, and a LLaMA~\cite{touvron2023llama} language model as the backbone. Given an image and a text prompt, the encoder and projector produce a sequence of visual tokens $T^{v} = \{T^{v}_{1}, T^{v}_{2}, ...T^{v}_{N} \}$, while the prompt is tokenized into $T^{l} = \{T^{l}_{1}, T^{l}_{2}, ..., T^{l}_{M} \}$. These tokens are concatenated and passed to the language model, which autoregressively generates actions as control outputs. However, the large number of visual tokens, combined with highly repetitive action outputs, leads to significant computational overhead. These observations highlight two key sources of inefficiency in VLA models: visual token redundancy and temporal redundancy in action generation. In the following sections, we introduce methods to address both aspects and improve inference efficiency without additional training.

\subsection{Visual Token Selection Strategy via Information Contribution Theory}
\label{Visual Token Selection Strategy via Information Contribution Theory}



We observe that VLA models exhibit visual token redundancy patterns similar to those found in VLMs~\cite{chen2024image}, where attention distributions are highly sparse beyond the initial few transformer layers (see Appendix~\ref{sec:app_Visual_Redundancy}). This motivates our token selection strategy based on information contribution theory, which identifies tokens that best preserve the structure of the visual feature space. However, most recent architectures (e.g., OpenVLA) rely on Flash Attention~\cite{dao2022flashattention}, making traditional attention score-based selection infeasible. To address this, we directly operate on the attention output matrix and rank tokens by their estimated contribution to the global feature representation.


Let $\hat{T}^{v} \in \mathbb{R}^{N \times d}$ denote the attention output matrix corresponding to the visual tokens, where $N$ is the number of tokens and $d$ is the hidden dimension. To quantify the amount of information contained in $\hat{T}^{v}$, we perform singular value decomposition (SVD), where $\hat{T}^{v} = U \Sigma V^\top$. Here, $U \in \mathbb{R}^{N \times N}$ contains the left singular vectors, $V \in \mathbb{R}^{d \times d}$ contains the right singular vectors, and $\Sigma \in \mathbb{R}^{N \times d}$ is a diagonal matrix of singular values $\sigma_i$ sorted in descending order. Each token representation $\hat{T}^{v}(x)$ corresponds to a row of $\hat{T}^{v}$ and can be expressed as:

\vspace{-0.2cm}
\begin{equation}
    \hat{T}^{v}(x) = \sum_{i=1}^{r} u_{xi} \sigma_i v_i^\top
\end{equation}

where $r$ is the effective rank of the matrix. We define the \textit{information contribution score} (ICS) of the $x$-th token as:

\vspace{-0.2cm}
\begin{equation}
    C(x) = \sum_{i=1}^{r} \left| u_{xi} \sigma_i \right|
\end{equation}
\vspace{-0.2cm}

This score measures the magnitude of the token's projection onto the dominant singular directions, weighted by their corresponding singular values. Tokens with higher $C(x)$ values are expected to contribute more significantly to the overall representation.


    

\begin{wrapfigure}{h}{6cm}
    \vspace{-0.7cm}
    \centering
    \includegraphics[width=0.45\textwidth]{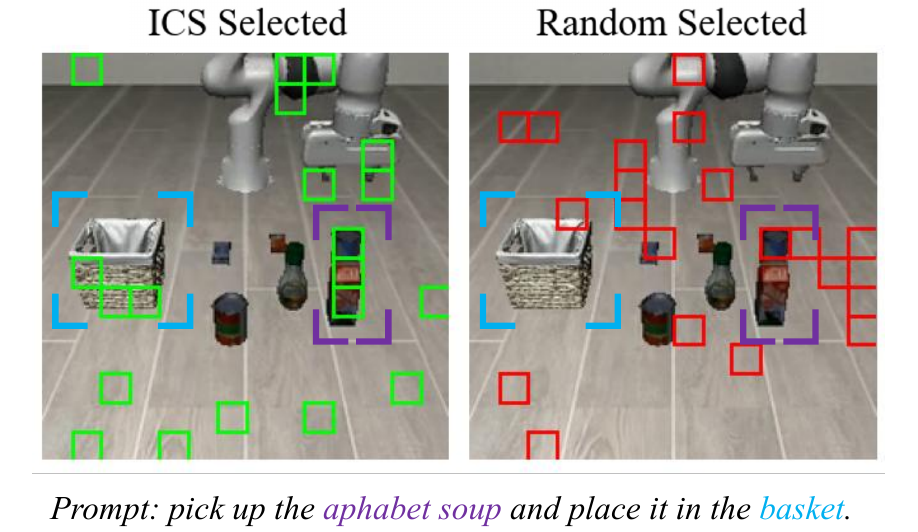}
    \vspace{-0.5cm}
    \caption{Comparison of visual token selection strategies on a sample image. 
Left: patches selected using the proposed ICS. 
Right: patches selected uniformly at random. 
Patches selected by ICS tend to focus on semantically meaningful and information-dense regions. } 
    \vspace{-0.5cm}
    \label{fig2_ICS_Theory}
\end{wrapfigure}

\paragraph{Theoretical Justification.}
To theoretically justify the superiority of ICS-based token selection over random sampling, we analyze the information retention in the top-$K$ selected tokens. Let $S \subset \{1, \dots, N\}$ with $K\subset (1,  N)$ be the number of selected token indices, and let $T_S^v \in \mathbb{R}^{K \times d}$ be the corresponding token matrix. The retained information is measured by the Frobenius norm of its projection onto the top-$r$ singular directions:

\vspace{-0.2cm}
\begin{equation}
    \mathcal{I}(S) = \left\| T_S^v V_r \right\|_F^2 = \sum_{x \in S} \sum_{i=1}^{r} (u_{xi} \sigma_i)^2
\end{equation}

Maximizing $\mathcal{I}(S)$ ensures the preservation of the dominant subspace of $\hat{T}^{v}$. Although $C(x)$ is defined via absolute values, it provides a greedy approximation to maximizing $\mathcal{I}(S)$. Specifically, by the Cauchy–Schwarz inequality:

\vspace{-0.2cm}
\begin{equation}
    C(x)^2 \leq r \sum_{i=1}^{r} (u_{xi} \sigma_i)^2
\end{equation}
\vspace{-0.2cm}

Thus, ranking tokens by $C(x)$ identifies those with high energy in the top singular directions.The retained information of the top-$K$ tokens is:

\vspace{-0.2cm}
\begin{equation}
    \mathcal{I}(S_{C}) = \sum_{x \in S_{C}} \sum_{i=1}^{k} (u_{xi} \sigma_i)^2
\end{equation}
\vspace{-0.2cm}

In contrast, for uniformly random selection, the expected information retained is:

\vspace{-0.2cm}
\begin{equation}
    \mathbb{E}[\mathcal{I}(S_{\text{rand}})] = \frac{K}{N} \sum_{x=1}^{N} \sum_{i=1}^{r} (u_{xi} \sigma_i)^2
\end{equation}
\vspace{-0.2cm}

Since the top-$K$ tokens ranked by $C(x)$ dominate this global sum, we have $\mathcal{I}(S_{C}) \geq \mathbb{E}[\mathcal{I}(S_{\text{rand}})]$. Fig.~\ref{fig2_ICS_Theory} illustrates this phenomenon by showing that tokens selected via ICS tend to concentrate on information-rich regions and achieve higher information retention than randomly selected tokens. Therefore, the ICS-based selection strategy retains more structural information in practice.

\subsection{Token-Aware Action Reuse Strategy}

\label{Token-Aware Action reuse strategy}
Action outputs in VLA models often exhibit minimal variation or remain identical across frames (Fig.\ref{fig1_Motivation}), and can be regarded as redundant actions suitable for direct reuse to accelerate inference. As shown in Fig.\ref{fig2_Framework}, FlashTrigger decides whether to reuse the previous action or perform a new pruned inference. It consists of Action Memory, Token Memory, and a Trigger Thinking module. To ensure stability, reuse is not applied in the first two frames; the current frame is denoted as the $s$-th with $s > 2$. At the action level, consistency is measured by the variation in action vectors $\vec{A}$. Action Memory stores the outputs from the previous two frames, $\vec{A}(s-2)$ and $\vec{A}(s-1)$, and computes the angle $\alpha$ between them to quantify the change:

\vspace{-0.2cm}
\begin{equation}
    \alpha(s) = Arccos(\frac{\vec{A}(s-2)\cdot \vec{A}(s-1)}{||\vec{A}(s-2)||\times ||\vec{A}(s-1)|| })
\end{equation}
\vspace{-0.2cm}

In token level, the change of set $I$ computed in Section~\ref{Visual Token Selection Strategy via Information Contribution Theory} is used to determine the degree of change in environment with a VLA model perspective. The Token Memory dynamically updates and stores the computed $I$ from the previous two frames: $I(s-2)$ and $I(s-1)$. We calculate the intersection ratio $\phi$ between $I(s-2)$ and $I(s-1)$ to determine the extent of the change of $I$:

\vspace{-0.2cm}
\begin{equation}
    \phi(s) = \frac{Len\left[ I(s-2) \cap I(s-1)\right] }{Len\left[I(s-1) \right]} 
\end{equation}
where $Len(\cdot)$ returns the number of elements in the set. 

At the same time, the lower limit of $\alpha(s)$ change is denoted by $\varepsilon _{1}$. We define $\delta$ as the maximum number of allowed changes in the set of visual tokens at the two action steps before and after. Thus the lower limit of the threshold $\varepsilon_{2}$ at which $\phi(s)$ is allowed to change can be defined:

\vspace{-0.2cm}
\begin{equation}
    \varepsilon _{2} = 1-\frac{\delta}{Len\left[I(s-1) \right]}
\label{eq:Th2}
\end{equation}
\vspace{-0.1cm}
Before each inference, the Trigger Thinking module thinks ahead about whether this inference should directly reuse $\vec{A}(s-1)$ or recalculate how to act. It can be represented as:
\begin{equation}
Trigger \ Thinking = \left\{
\begin{aligned}
&Reuse \ Action, \ &if \ \alpha(s) \textgreater \varepsilon _{1} \ and \ \phi(s) \textgreater \varepsilon _{2} \\
&Pruned \ Inference, \ &else\\
\end{aligned}
\right.
\end{equation}
\vspace{-0.2cm}

When the reuse condition is not met, VLA performs pruned inference using the informative visual token set $I$ instead of the full token set. In other words, \methodshort{} consistently accelerates the inference process, regardless of whether the action is reused.

%% file: sec/4_Experiments.tex
\section{Experiment}
\label{sec:Experiment}
\subsection{Experimental Setup}
\label{sec:Experiment_Setup}
\paragraph{Evaluation Environment.}
We evaluate the performance of \methodshort{} on the LIBERO simulation benchmark, which features a simulated Franka Emika Panda robotic arm and provides multimodal demonstration data, including camera observations, robot states, task labels, and delta end-effector pose actions. The benchmark consists of four task suites—LIBERO-Spatial, LIBERO-Object, LIBERO-Goal, and LIBERO-Long—each comprising 500 expert demonstrations across 10 tasks. These suites are designed to test policy generalization under varying spatial layouts, object types, goal specifications, and long-horizon task sequences.
\paragraph{Implementation Details.}
We apply the \methodshort{} framework to accelerate the OpenVLA model fine-tuned on the LIBERO simulation benchmark. All experiments are conducted on a single NVIDIA H100 GPU. We report three evaluation metrics under different visual token configurations: success rate (SR), inference latency, and visual-token FLOPs. Latency is measured via wall-clock time, and FLOPs are computed following the method described in Appendix~\ref{sec:app_Computation_Cost_Estimation}. To assess real-world speedup, we utilize the \texttt{torch.profiler} tool for runtime profiling.
\methodshort{} includes a threshold-based action reuse mechanism whose sensitivity is governed by two hyperparameters, $\varepsilon_1$ and $\varepsilon_2$. Unless otherwise specified, we fix $\varepsilon_1 = 2$ and assign $\delta$ values based on the number of visual tokens as follows: ${(192, 3),\ (160, 4.5),\ (128, 5),\ (96, 5.5)}$. The $\delta$ values are converted to $\varepsilon_2$ using Equation~\ref{eq:Th2}. A detailed hyperparameter sensitivity study is presented in Section~\ref{sec:Experiment_Ablation_Study}.

\renewcommand{\arraystretch}{1} 
\begin{table}[t]
\caption{
Main results of \methodshort{} under different visual token budgets across four task suites in the LIBERO benchmark.
We report success rate (SR), visual-token FLOPs, and inference latency at five token settings.
With 160 visual tokens, \methodshort{} achieves a strong balance between SR and efficiency—reducing FLOPs by 55.7\% and latency by 36.0\%—while maintaining comparable or even improved success rates across most tasks.
}

\renewcommand{\arraystretch}{0.5} 
\label{tab:Main_Results}
\centering 
\resizebox{1\textwidth}{!}{
{\fontsize{9pt}{10}\selectfont
\begin{tabular}{cc|ccccc}
\toprule
\multicolumn{2}{c|}{\textbf{Task / Visual token}}           & \textbf{\begin{tabular}[c]{@{}c@{}}256\\ (Baseline)\end{tabular}} & \textbf{\begin{tabular}[c]{@{}c@{}}192\\ (75\%)\end{tabular}} & \textbf{\begin{tabular}[c]{@{}c@{}}160\\ (62.5\%)\end{tabular}} & \textbf{\begin{tabular}[c]{@{}c@{}}128\\ (50\%)\end{tabular}} & \textbf{\begin{tabular}[c]{@{}c@{}}96\\ (37.5\%)\end{tabular}} \\ \midrule
\multicolumn{1}{c|}{}                        & SR (\%)      & 84.2                                                              & 81.8 (\textcolor{DeepGreen}{$-$ 2.4})                                                         & 82.6 (\textcolor{DeepGreen}{$-$ 1.6})                                                           & 75.4 (\textcolor{DeepGreen}{$-$ 8.8})                                                          & 67 (\textcolor{DeepGreen}{$-$ 17.2})                                                            \\
\multicolumn{1}{c|}{\textbf{LIBERO-Spatial}} & Flops (\(\scalebox{0.8}{10}^{\scalebox{0.6}{12}}\))    & 1.31                                                              & 0.8 (\textcolor{red}{$\downarrow$ 38.9\%})                                                          & 0.66 (\textcolor{red}{$\downarrow$ 49.6\%})                                                           & 0.51 (\textcolor{red}{$\downarrow$ 61.1\%})                                                         & 0.43 (\textcolor{red}{$\downarrow$ 67.2\%})                                                          \\
\multicolumn{1}{c|}{}                        & Latency (ms) & 82.7                                                              & 62.7 (\textcolor{red}{$\downarrow$ 24.2\%})                                                          & 61.2 (\textcolor{red}{$\downarrow$ 26.0\%})                                                            & 58.1 (\textcolor{red}{$\downarrow$ 29.7\%})                                                          & 58.9 (\textcolor{red}{$\downarrow$ 28.8\%})                                                          \\ \midrule
\multicolumn{1}{c|}{}                        & SR (\%)      & 86.4                                                              & 86.6 (\textcolor{red}{$+$ 0.2})                                                         & 86.6 (\textcolor{red}{$+$ 0.2})                                                           & 85.2 (\textcolor{DeepGreen}{$-$ 1.2})                                                         & 83.6 (\textcolor{DeepGreen}{$-$ 2.8})                                                          \\
\multicolumn{1}{c|}{\textbf{LIBERO-Object}}  & Flops (\(\scalebox{0.8}{10}^{\scalebox{0.6}{12}}\))    & 1.31                                                              & 0.74 (\textcolor{red}{$\downarrow$ 43.5\%})                                                         & 0.57 (\textcolor{red}{$\downarrow$ 56.5\%})                                                           & 0.42 (\textcolor{red}{$\downarrow$ 67.9\%})                                                         & 0.33 (\textcolor{red}{$\downarrow$ 74.8\%})                                                          \\
\multicolumn{1}{c|}{}                        & Latency (ms) & 82.7                                                              & 58.8 (\textcolor{red}{$\downarrow$ 28.9\%})                                                         & 53.1 (\textcolor{red}{$\downarrow$ 35.8\%})                                                           & 45.3 (\textcolor{red}{$\downarrow$ 45.2\%})                                                         & 47.2 (\textcolor{red}{$\downarrow$ 42.9\%})                                                          \\ \midrule
\multicolumn{1}{c|}{}                        & SR (\%)      & 75.4                                                              & 76.2 (\textcolor{red}{$+$ 0.8})                                                         & 78.8 (\textcolor{red}{$+$ 3.4})                                                           & 76.8  (\textcolor{red}{$+$ 1.4})                                                        & 70.2 (\textcolor{DeepGreen}{$-$ 5.2})                                                          \\
\multicolumn{1}{c|}{\textbf{LIBERO-Goal}}    & Flops (\(\scalebox{0.8}{10}^{\scalebox{0.6}{12}}\))    & 1.31                                                              & 0.71 (\textcolor{red}{$\downarrow$ 45.8\%})                                                         & 0.6 (\textcolor{red}{$\downarrow$ 54.2\%})                                                            & 0.49 (\textcolor{red}{$\downarrow$ 62.6\%})                                                         & 0.37 (\textcolor{red}{$\downarrow$ 71.8\%})                                                          \\
\multicolumn{1}{c|}{}                        & Latency (ms) & 82.7                                                              & 55.2 (\textcolor{red}{$\downarrow$ 33.3\%})                                                         & 56.8 (\textcolor{red}{$\downarrow$ 31.3\%})                                                           & 54.1 (\textcolor{red}{$\downarrow$ 34.6\%})                                                         & 53.8 (\textcolor{red}{$\downarrow$ 34.9\%})                                                          \\ \midrule
\multicolumn{1}{c|}{}                        & SR (\%)      & 51.4                                                              & 50.2 (\textcolor{DeepGreen}{$-$ 1.2})                                                         & 46.8 (\textcolor{DeepGreen}{$-$ 4.6})                                                           & 46.4 (\textcolor{DeepGreen}{$-$ 5.0})                                                         & 45.2 (\textcolor{DeepGreen}{$-$ 6.2})                                                          \\
\multicolumn{1}{c|}{\textbf{LIBERO-Long}}    & Flops (\(\scalebox{0.8}{10}^{\scalebox{0.6}{12}}\))    & 1.31                                                              & 0.54 (\textcolor{red}{$\downarrow$ 58.8\%})                                                         & 0.47 (\textcolor{red}{$\downarrow$ 64.1\%})                                                           & 0.4 (\textcolor{red}{$\downarrow$ 69.5\%})                                                          & 0.32 (\textcolor{red}{$\downarrow$ 75.6\%})                                                          \\
\multicolumn{1}{c|}{}                        & Latency (ms) & 82.7                                                              & 42.43 (\textcolor{red}{$\downarrow$ 48.7\%})                                                        & 40.35 (\textcolor{red}{$\downarrow$ 51.2\%})                                                          & 38.31 (\textcolor{red}{$\downarrow$ 53.7\%})                                                        & 44.05 (\textcolor{red}{$\downarrow$ 46.7\%})                                                         \\ \midrule
\multicolumn{1}{c|}{}                        & SR (\%)      & 74.4                                                              & 73.7  (\textcolor{DeepGreen}{$-$ 0.7})                                                        & 73.7 (\textcolor{DeepGreen}{$-$ 0.7})                                                           & 71.0 (\textcolor{DeepGreen}{$-$ 3.4})                                                         & 66.5 (\textcolor{DeepGreen}{$-$ 7.9})                                                          \\
\multicolumn{1}{c|}{\textbf{Average}}        & Flops (\(\scalebox{0.8}{10}^{\scalebox{0.6}{12}}\))    & 1.31                                                              & 0.70 (\textcolor{red}{$\downarrow$ 46.6\%})                                                         & 0.58 (\textcolor{red}{$\downarrow$ 55.7\%})                                                           & 0.46 (\textcolor{red}{$\downarrow$ 64.9\%})                                                         & 0.36 (\textcolor{red}{$\downarrow$ 72.5\%})                                                          \\
\multicolumn{1}{c|}{}                        & Latency (ms) & 82.7                                                              & 54.8 (\textcolor{red}{$\downarrow$ 33.7\%})                                                         & 52.9 (\textcolor{red}{$\downarrow$ 36.0\%})                                                          & 49.0 (\textcolor{red}{$\downarrow$ 40.7\%})                                                         & 51.0 (\textcolor{red}{$\downarrow$ 38.3\%})                                                          \\ \bottomrule
\end{tabular}
}
}
\vspace{-0.3cm}
\end{table}

\subsection{Main Results on LIBERO Benchmark}
\label{sec:Experiment_Main_Results}
We evaluate the performance of \methodshort{} across four task suites in the LIBERO benchmark under varying visual token budgets. As shown in Table~\ref{tab:Main_Results}, the 160-token configuration (62.5\% of the original) offers the best trade-off between accuracy and efficiency. Compared to the full 256-token baseline, it reduces visual-token FLOPs by 55.7\% (from 1.31 to 0.58 $\times 10^{12}$) and lowers inference latency by 36.0\% (from 82.7ms to 52.9ms), while maintaining the same average success rate (73.7\% vs. 74.4\%).
Interestingly, we observe a slight performance gain at 160 tokens in both \textbf{LIBERO-Object} and \textbf{LIBERO-Goal}, where the success rate increases from 86.4\% to 86.6\% and from 75.4\% to 78.8\%, respectively. This suggests that modest token pruning may even help mitigate redundancy and stabilize policy execution. In contrast, \textbf{LIBERO-Long} is more sensitive to pruning, showing a larger SR decline when the token count drops below 160. Nevertheless, even in such cases, \methodshort{} achieves substantial savings in computational cost (e.g., 64.1\% FLOPs reduction at 160 tokens).
These results highlight that \methodshort{}, particularly at the 160-token configuration, significantly improves inference efficiency without compromising performance, making it suitable for real-world deployment across a wide range of embodied tasks.
Further, we visualize the trajectories of FlashVLA and the baseline during successful task executions (Fig.~\ref{fig5_Trajectory}), showing that FlashVLA exhibits smoother and more stable motion while achieving comparable outcomes.

\begin{figure*}
    \centering
    \includegraphics[width=0.95\linewidth]{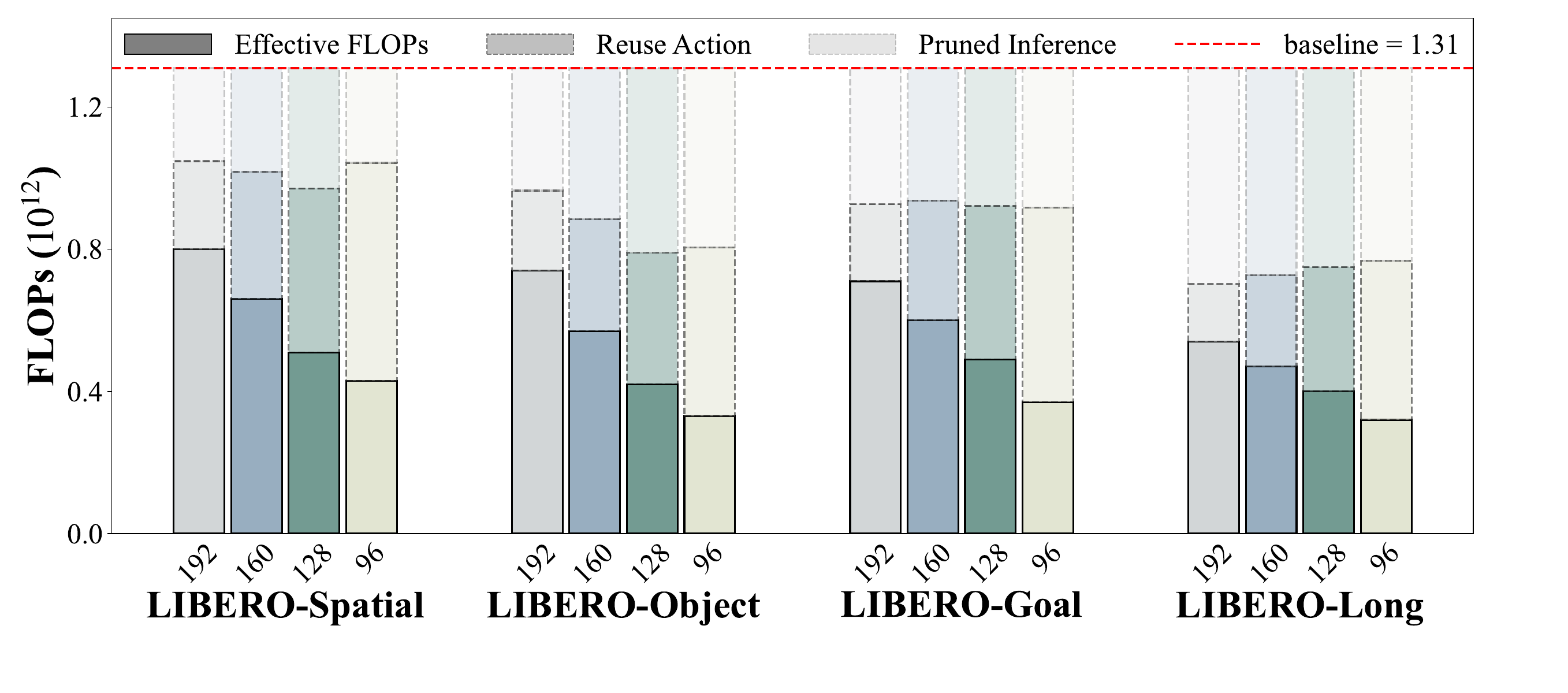}
    \vspace{-0.4cm}
    \caption{FLOPs breakdown of FlashVLA across four LIBERO tasks under different visual token budgets. Each bar shows the cumulative reduction in FLOPs contributed by token pruning and computation reuse. FlashVLA consistently operates below the baseline FLOPs (dashed line), demonstrating the effectiveness of the dual-path acceleration strategy.}
    \vspace{-0.2cm}
    \label{fig5_FLOPs_Breakdown}
\end{figure*}

\subsection{Ablation Study}
\label{sec:Experiment_Ablation_Study}
\paragraph{Effect of Token Pruning and Action Reuse.}
To assess the contribution of each component in \methodshort{}, we conduct ablation studies on the LIBERO-Spatial suite by disabling \textit{Pruned Inference} and \textit{Action Reuse} individually. As shown in Table~\ref{tab:Ablation_Study_1}, both modules are critical for efficient inference with minimal performance loss. Removing Action Reuse (w/o Action Reuse) maintains the benefit of token pruning and achieves significant FLOPs reduction (e.g., 0.66 at 160 tokens vs. 0.85 without reuse), while maintaining comparable success rate. This indicates that Action Reuse primarily improves efficiency rather than accuracy. In contrast, disabling Pruned Inference (w/o Pruned Inference) results in consistently higher FLOPs across all token settings, with only marginal SR improvement—for example, at 128 tokens, FLOPs increase from 0.51 (FlashVLA) to 0.97 ($\times 10^{12}$), while SR improves from 75.4\% to 81.0\%. The full FlashVLA configuration combines both strategies and achieves the best trade-off between performance and efficiency. Fig.~\ref{fig5_FLOPs_Breakdown} further demonstrates that the combination of both mechanisms effectively reduces the computational cost of VLA models.

\paragraph{Component Analysis of Action Reuse Module.}
To evaluate the internal design of the Action Reuse module in \methodshort{}, we conduct a component-level ablation on the LIBERO-Spatial suite by removing either the action vector signal (\textit{w/o ActionVector}) or the token information signal (\textit{w/o TokenSet}). As shown in Table~\ref{tab:Ablation_Study_1}, disabling either component degrades overall performance. Without the action vector, the success rate drops notably under tight token budgets (e.g., 82.6\% $\rightarrow$ 65.4\% at 96 tokens), indicating that temporal consistency in action space is crucial for reliable reuse. Without token stability, the model tends to reuse too aggressively, achieving the lowest FLOPs (e.g., 0.29 at 96 tokens) but unstable control (e.g., SR drops to 62.2\%). These results highlight the complementary roles of both signals: the action vector captures temporal continuity, while token stability reflects environmental change. Together, they enable the reuse module to balance efficiency and robustness.

\renewcommand{\arraystretch}{1.4} 
\begin{table}[t]
\caption{We examine the effects of removing the two core modules—Pruned Inference and Action Reuse—as well as the impact of disabling the \textit{ActionVector} and \textit{TokenSet} components within the reuse mechanism. FlashVLA consistently achieves the best balance between accuracy and efficiency, while removing either module or component leads to higher FLOPs or reduced performance.}
\label{tab:Ablation_Study_1}
\centering
\resizebox{\textwidth}{!}{%
\begin{tabular}{c| cc cc cc cc cc}
\toprule
\multirow{2}{*}{\textbf{Method}} 
& \multicolumn{2}{c}{\textbf{256 Tokens}} & \multicolumn{2}{c}{\textbf{192 Tokens}} 
& \multicolumn{2}{c}{\textbf{160 Tokens}} & \multicolumn{2}{c}{\textbf{128 Tokens}} & \multicolumn{2}{c}{\textbf{96 Tokens}} \\
\cline{2-11}
& \makecell{\rule{0pt}{10pt}SR\\(\%)} 
& \makecell{\rule{0pt}{10pt}FLOPs\\(\(\scalebox{0.8}{1\!0}^{\scalebox{0.6}{12}}\))} 
& \makecell{\rule{0pt}{10pt}SR\\(\%)} 
& \makecell{\rule{0pt}{10pt}FLOPs\\(\(\scalebox{0.8}{1\!0}^{\scalebox{0.6}{12}}\))} 
& \makecell{\rule{0pt}{10pt}SR\\(\%)} 
& \makecell{\rule{0pt}{10pt}FLOPs\\(\(\scalebox{0.8}{1\!0}^{\scalebox{0.6}{12}}\))} 
& \makecell{\rule{0pt}{10pt}SR\\(\%)} 
& \makecell{\rule{0pt}{10pt}FLOPs\\(\(\scalebox{0.8}{1\!0}^{\scalebox{0.6}{12}}\))}  
& \makecell{\rule{0pt}{10pt}SR\\(\%)} 
& \makecell{\rule{0pt}{10pt}FLOPs\\(\(\scalebox{0.8}{1\!0}^{\scalebox{0.6}{12}}\))}    \\
\midrule
\textbf{w/o Action Reuse}     & \makecell{84.2} & \makecell{1.31} & \makecell{84.6} & \makecell{1.00} & \makecell{83.6} & \makecell{0.85} & \makecell{78.2} & \makecell{0.69} & \makecell{67.8} & \makecell{0.54} \\
\textbf{w/o Pruned Inference} & \makecell{84.2} & \makecell{1.31} & \makecell{82.2} & \makecell{1.04} & \makecell{80.6} & \makecell{1.01} & \makecell{81.0} & \makecell{0.97} & \makecell{80.2} & \makecell{0.97} \\
\textbf{w/o ActionVector for Action Reuse} & \makecell{84.2} & \makecell{1.31} & \makecell{81.6} & \makecell{0.68} & \makecell{79.6} & \makecell{0.56} & \makecell{73.2} & \makecell{0.44} & \makecell{65.4} & \makecell{0.35} \\
\textbf{w/o TokenSet for Action Reuse} & \makecell{84.2} & \makecell{1.31} & \makecell{76.4} & \makecell{0.52} & \makecell{74.8} & \makecell{0.44} & \makecell{73.4} & \makecell{0.37} & \makecell{62.2} & \makecell{0.29} \\
\midrule
\textbf{FlashVLA}                        & \makecell{84.2} & \makecell{1.31} & \makecell{81.8} & \makecell{0.80} & \makecell{82.6} & \makecell{0.66} & \makecell{75.4} & \makecell{0.51} & \makecell{67.0} & \makecell{0.43} \\
\bottomrule
\end{tabular}
}
\vspace{-0.2cm}
\end{table}

\begin{figure}
\begin{minipage}[t]{0.52\textwidth}
    \centering
    \includegraphics[scale=0.55]{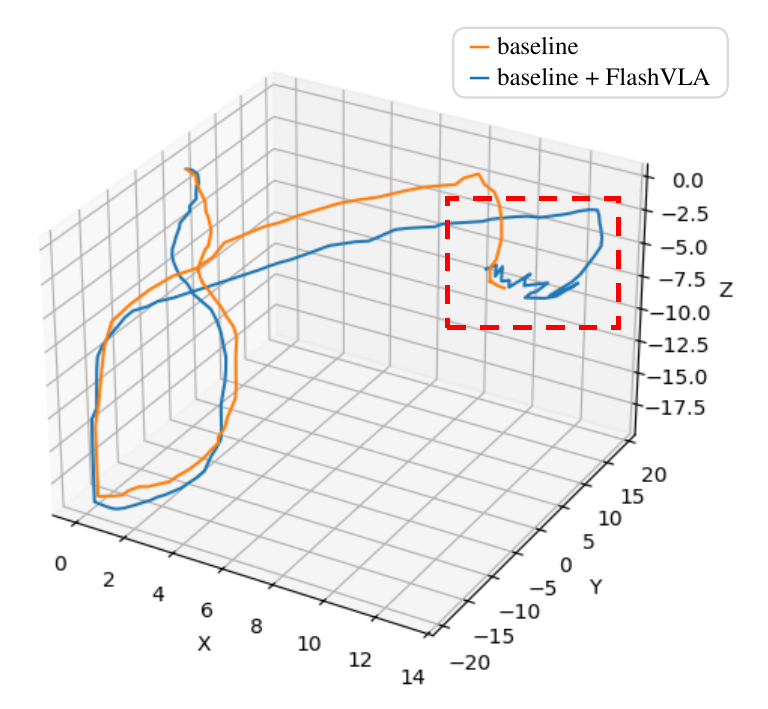}
    \caption{Trajectory of action. We visualize the trajectory of action in 3-dimensional space. The location of red dashed box illustrates the smoother trajectory of \methodshort{} for the same task.}
    \label{fig5_Trajectory}
\end{minipage}
\hspace{0.015\textwidth}
\begin{minipage}[t]{0.42\textwidth}
    \centering
    \includegraphics[scale=0.9]{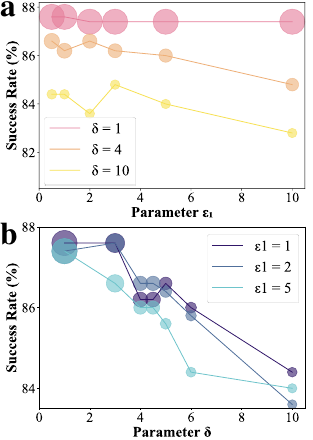}
    \caption{Hyperparameter Sensitivity. Experiments are conducted to investigate the effect of parameter $\varepsilon_1$ and  $\delta$. The size of point  represents the size of the FLOPs.}
    \label{fig6_hyperparameter}
\end{minipage} 
\end{figure}

\paragraph{Hyperparameter Sensitivity.}
We conduct experiments to analyze the impact of two hyperparameters, $\varepsilon_1$ and $\delta$, on performance and efficiency, as shown in Fig.\ref{fig6_hyperparameter}. Fig.\ref{fig6_hyperparameter}(a) shows that varying $\varepsilon_1$ has negligible effect on both success rate and FLOPs, indicating that our method is largely insensitive to this parameter. In contrast, Fig.~\ref{fig6_hyperparameter}(b) shows that $\delta$ significantly affects both metrics: increasing $\delta$ reduces computation cost but leads to a drop in performance. This provides flexibility to trade-off accuracy and efficiency based on application needs.

\setlength{\aboverulesep}{0.3pt}
\setlength{\aboverulesep}{0.3pt}
\setlength{\aboverulesep}{0.3pt}
\setlength{\aboverulesep}{0.3pt}
\begin{table}[t]
\vspace{-0.2cm}
\caption{Comparison between \methodshort{} and FastV on the LIBERO-Object task suite under various visual token budgets.}
\label{tab:Token_Pruning_Method}
\centering
\resizebox{1\textwidth}{!}{
{
\fontsize{7pt}{6}\selectfont  
\begin{tabular}{cc|ccccc}
\toprule
\multicolumn{2}{c|}{\textbf{Method / Visual token}} & 
\textbf{\begin{tabular}[c]{@{}c@{}}256\\ (Baseline)\end{tabular}} & 
\textbf{\begin{tabular}[c]{@{}c@{}}192\\ (75\%)\end{tabular}} & 
\textbf{\begin{tabular}[c]{@{}c@{}}160\\ (62.5\%)\end{tabular}} & 
\textbf{\begin{tabular}[c]{@{}c@{}}128\\ (50\%)\end{tabular}} & 
\textbf{\begin{tabular}[c]{@{}c@{}}96\\ (37.5\%)\end{tabular}} \\ 
\midrule

\multicolumn{1}{c|}{} & SR (\%) & 86.4 & 86.2 (\textcolor{DeepGreen}{$-$ 0.2})  & 84.8 (\textcolor{DeepGreen}{$-$ 1.6})   & 85.2 (\textcolor{DeepGreen}{$-$ 1.2})  & 85.2 (\textcolor{DeepGreen}{$-$ 1.2})  \\
\multicolumn{1}{c|}{\textbf{FastV~\cite{chen2024image}}} & Flops (\(\scalebox{0.8}{10}^{\scalebox{0.6}{12}}\)) & 1.31 & 1.00 (\textcolor{red}{$\downarrow$ 23.7\%})  & 0.85 (\textcolor{red}{$\downarrow$ 35.1\%})  & 0.69 (\textcolor{red}{$\downarrow$ 47.3\%})  & 0.54 (\textcolor{red}{$\downarrow$ 58.8\%})  \\
\multicolumn{1}{c|}{} & Latency (ms) & 82.7 & 81.3 (\textcolor{red}{$\downarrow$ 1.7\%})  & 79.9 (\textcolor{red}{$\downarrow$ 3.4\%})  & 78.9 (\textcolor{red}{$\downarrow$ 4.6\%})  & 77.8 (\textcolor{red}{$\downarrow$ 5.9\%})  \\
\midrule

\multicolumn{1}{c|}{} & SR (\%) & 86.4 & 86.6 (\textcolor{red}{$+$ 0.2})  & 86.6 (\textcolor{red}{$+$ 0.2})  & 85.2 (\textcolor{DeepGreen}{$-$ 0.8})  & 83.6 (\textcolor{DeepGreen}{$-$ 2.8})  \\
\multicolumn{1}{c|}{\textbf{FlashVLA}} & Flops (\(\scalebox{0.8}{10}^{\scalebox{0.6}{12}}\)) & 1.31 & 0.74 (\textcolor{red}{$\downarrow$ 51.1\%}) & 0.57 (\textcolor{red}{$\downarrow$ 56.5\%}) & 0.42 (\textcolor{red}{$\downarrow$ 67.3\%}) & 0.33 (\textcolor{red}{$\downarrow$ 74.8\%}) \\
\multicolumn{1}{c|}{} & Latency (ms) & 82.7 & 58.8 (\textcolor{red}{$\downarrow$ 28.9\%}) & 53.1 (\textcolor{red}{$\downarrow$ 35.8\%}) & 45.3 (\textcolor{red}{$\downarrow$ 45.2\%}) & 47.2 (\textcolor{red}{$\downarrow$ 42.9\%}) \\
\bottomrule
\end{tabular}

}
}
\end{table}

\subsection{Comparison with Token Pruning Methods}

\label{sec:Experiment_Token_Pruning_Method}
\methodshort{} is compared with FastV~\cite{chen2024image}, a \methodshort{} is compared with FastV~\cite{chen2024image}, a representative dynamic token pruning baseline on the LIBERO-Object suite under varying visual token budgets. As shown in Table~\ref{tab:Token_Pruning_Method}, \methodshort{} consistently achieves comparable or better performance with lower FLOPs. At 160 tokens, for example, it attains a higher success rate (86.6\% vs. 84.8\%) while reducing FLOPs from 0.85 to 0.57 ($\times 10^{12}$).
While both methods fall under the dynamic token pruning paradigm, a key advantage of \methodshort{} is compatibility with FlashAttention~\cite{dao2022flashattention}, a memory-efficient and GPU-optimized attention kernel widely used in LLMs. This enables FlashVLA to achieve lower memory overhead and faster inference, making it suitable for real-time deployment.

\begin{wraptable}{h}{7.5cm}
  \vspace{-0.5cm}
  \small
  \centering
  \renewcommand\tabcolsep{5.5pt}
  \renewcommand{\arraystretch}{0.95}
  \caption{Performance in VLAbench}
  \vspace{-0.2cm}
  \begin{tabular}{c|cc}
    \toprule
    Visual token  & SR(\%) & Flops (\(\scalebox{0.8}{10}^{\scalebox{0.6}{12}}\)) \\ \midrule
    256 (baseline) & 7.0     & 1.31         \\
    192           & 8.0 (\textcolor{red}{$+$ 1.0})      & 0.62 (\textcolor{red}{$\downarrow$ 52.7\%})        \\
    160           & 5.0 (\textcolor{DeepGreen}{$-$ 2.0})     & 0.62 (\textcolor{red}{$\downarrow$ 52.7\%})        \\
    128           & 6.0 (\textcolor{DeepGreen}{$-$ 1.0})     & 0.41 (\textcolor{red}{$\downarrow$ 68.7\%})        \\
    96            & 7.0 (\textcolor{red}{$+$ 0.0})     & 0.30 (\textcolor{red}{$\downarrow$ 77.1\%})        \\ \bottomrule
  \end{tabular}
  \label{tab:Other_Environmentd}
  \vspace{-0.4cm}
\end{wraptable}

\subsection{Generalization to Other Environments}

\label{sec:Other Environment}
To assess the generality of our approach, we evaluate it on VLAbench~\cite{zhang2024vlabench}, a simulated robot environment featuring more diverse and challenging tasks. Specifically, we test on the \textit{Select Painting} task using OpenVLA with LoRA~\cite{hu2022lora}-fine-tuned weights provided by the authors. As shown in Table~\ref{tab:Other_Environmentd}, although the overall success rate is low due to task difficulty, our method significantly reduces computational cost while preserving baseline performance.

%% file: sec/5_Conclusion.tex
\newpage
\vspace{-0.1cm}
\section{Conclusion}
\vspace{-0.1cm}
\label{sec:Conclusion}

We propose \methodshort{}, the first training-free and plug-and-play acceleration framework that enables action reuse in VLA models. By exploiting two forms of redundancy—temporal coherence across consecutive actions and visual token redundancy—\methodshort{} improves inference efficiency through token-aware action reuse and information-guided visual token pruning. Specifically, it reduces unnecessary computation both across action steps and within input tokens. Experiments on the LIBERO benchmark show that \methodshort{} reduces FLOPs by 55.7\% and latency by 36.0\%, with only a 0.7\% drop in success rate—demonstrating its practicality, effectiveness, and scalability for efficient VLA inference. In future work, we plan to explore additional inference acceleration techniques further tailored to the unique characteristics of VLA models.

%% file: sec/appendix.tex
\appendix
\onecolumn

\textbf{{\Large Appendix for \methodshort{}}}

\section{Algorithm flow of \methodshort{}}
\label{sec:app_algo}
This section provides a detailed description of the algorithmic workflow of \methodshort{}, as outlined in Algorithm~\ref{alg}. The overall execution is divided into two phases: initialization and iterative reasoning.

During the initialization phase (lines 1–5), the agent executes the first two steps without action reuse to establish initial context. For each of the first two frames, the model selects a subset of informative visual tokens using the strategy described in Section~\ref{Visual Token Selection Strategy via Information Contribution Theory}, performs pruned inference based on the selected tokens, executes the resulting action, and updates the \textit{Action Memory} and \textit{Token Memory} with the new observations and outputs.

The iterative phase begins thereafter (lines 6–19), and continues until the task is successfully completed. At each step, the \textbf{FlashTrigger} mechanism (Section~\ref{Token-Aware Action reuse strategy}) determines whether the previous action can be reused. If reuse is triggered and the last step did not already reuse an action, the model directly reuses the previous output and sets the reuse flag. Otherwise, the model selects a new visual token subset, runs pruned inference, updates both memories, and resets the reuse flag. Regardless of reuse, the action is executed in the environment and the task state is updated.

Once the task is complete, the loop exits.

\begin{algorithm}[ht]
    \caption{ \methodshort{} }
    \label{alg}

    \begin{algorithmic}[1]
    \fontsize{10pt}{13}\selectfont
        \For{\textit{i in range(2)}}
            \State  $\triangleright$ Select important visual token set, according to Section.~\ref{Visual Token Selection Strategy via Information Contribution Theory}
            \State $\triangleright$ Reasoning using these important visual token sets to get action
            \State $\triangleright$ Perform the action
            \State $\triangleright$ Update \textit{Action Memory} and \textit{Token Memory}, according to action and visual token set            
        \EndFor
        \While{\textit{Task State} \textbf{is} \textit{False}}
            \State $\triangleright$ Calculate flag \textit{Reuse Action}, according to \textbf{\textit{Flashtrigger}} in Section.~\ref{Token-Aware Action reuse strategy}
            \If{\textit{Reuse Action} \textbf{is} \textit{True} and \textit{last reuse} \textbf{is} \textit{False}}
                \State $\triangleright$ Reuse last action
                \State $\triangleright$ Set flag \textit{last reuse} as \textit{True}
            \Else
                \State  $\triangleright$ Select important visual token set, according to Section.~\ref{Visual Token Selection Strategy via Information Contribution Theory}
                \State $\triangleright$ Reasoning using these important visual token sets to get action
                \State $\triangleright$ Update \textit{Action Memory} and \textit{Token Memory}, according to action and visual token set
                \State $\triangleright$ Set flag \textit{last reuse} as \textit{False}
            \EndIf
            \State $\triangleright$ Perform the action            
            \State $\triangleright$ Update \textit{Task State}
            \If{\textit{Task State} \textbf{is} \textit{True}}
                \State $\triangleright$ \textbf{break}
            \EndIf
        \EndWhile
    \end{algorithmic}
    
\end{algorithm}

\section{Computation Cost Estimation}
\label{sec:app_Computation_Cost_Estimation}

To analyze the computational efficiency of FlashVLA, we estimate the FLOPs consumed by the Multi-Head Attention (MHA) and Feed-Forward Network (FFN) modules, which dominate the cost of Transformer-based architectures. In VLA models, visual tokens typically account for more than 80\% of the input, making them the primary contributor to overall inference cost. Since the number of language prompt tokens varies across tasks, we use the FLOPs associated with visual tokens as a consistent and representative measure of complexity.

The total FLOPs are estimated as:

\vspace{-0.4cm}
\begin{equation}
\label{eq:flops}
\text{FLOPs} = (1 - R) \times \left[ L_p \cdot (4nd^2 + 2n^2d + 2ndm) + (L - L_p) \cdot (4n_p d^2 + 2n_p^2 d + 2n_p d m) \right]
\end{equation}
\vspace{-0.3cm}

where:

\begin{itemize}
  \item $n$: total number of input tokens (visual + language),
  \item $d$: hidden dimension,
  \item $m$: intermediate dimension in the FFN module,
  \item $L$: total number of Transformer layers,
  \item $L_p$: layer index at which visual token pruning starts,
  \item $n_p$: number of visual tokens after pruning,
  \item $R$: action reuse rate.
\end{itemize}

By default, we set $L_p = 2$ during the prefill stage, meaning that full-token computation is retained in the first two layers to preserve early-layer representation quality. During decoding, FlashVLA reuses token selections from the prefill stage and sets $L_p = 0$. Therefore, the actual FLOPs of FlashVLA are slightly lower than the values reported in this paper, as both prefill and decoding stages are estimated using $L_p = 2$ for consistency.

\section{Visual Redundancy in VLA Models}
\label{sec:app_Visual_Redundancy}
\begin{figure*}
    \centering
    \includegraphics[width=\linewidth]{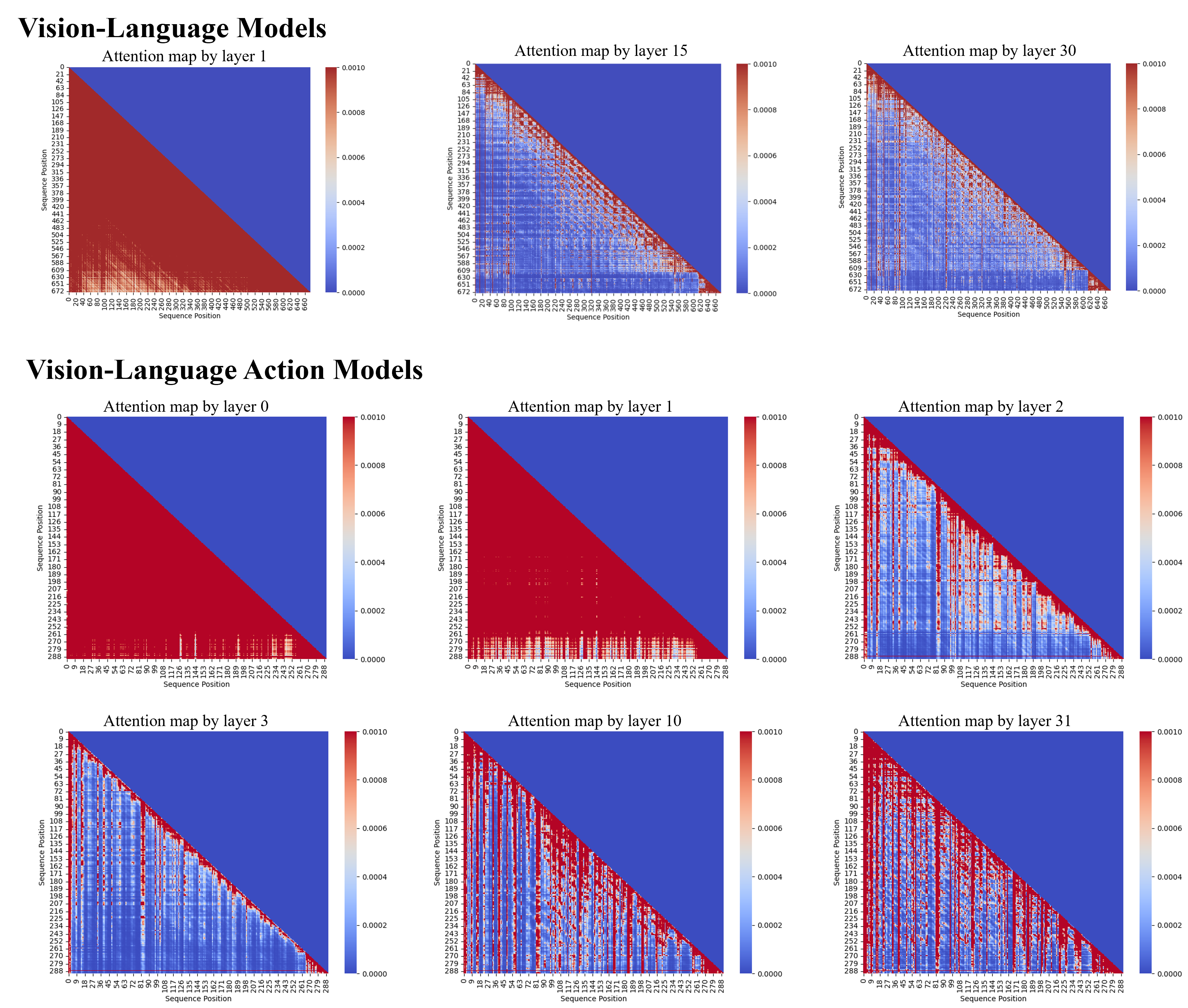}
    \caption{\textbf{Attention Map}: Layer-wise attention map visualizations in VLA and VLM models. Both models exhibit uniform attention distribution in the first layer, while attention becomes increasingly sparse from the second layer onward. This pattern suggests growing redundancy in token interactions, motivating token pruning strategies in deeper layers}
    \label{fig7_Attention_map}
\end{figure*}

\begin{figure*}
    \centering
    \includegraphics[width=\linewidth]{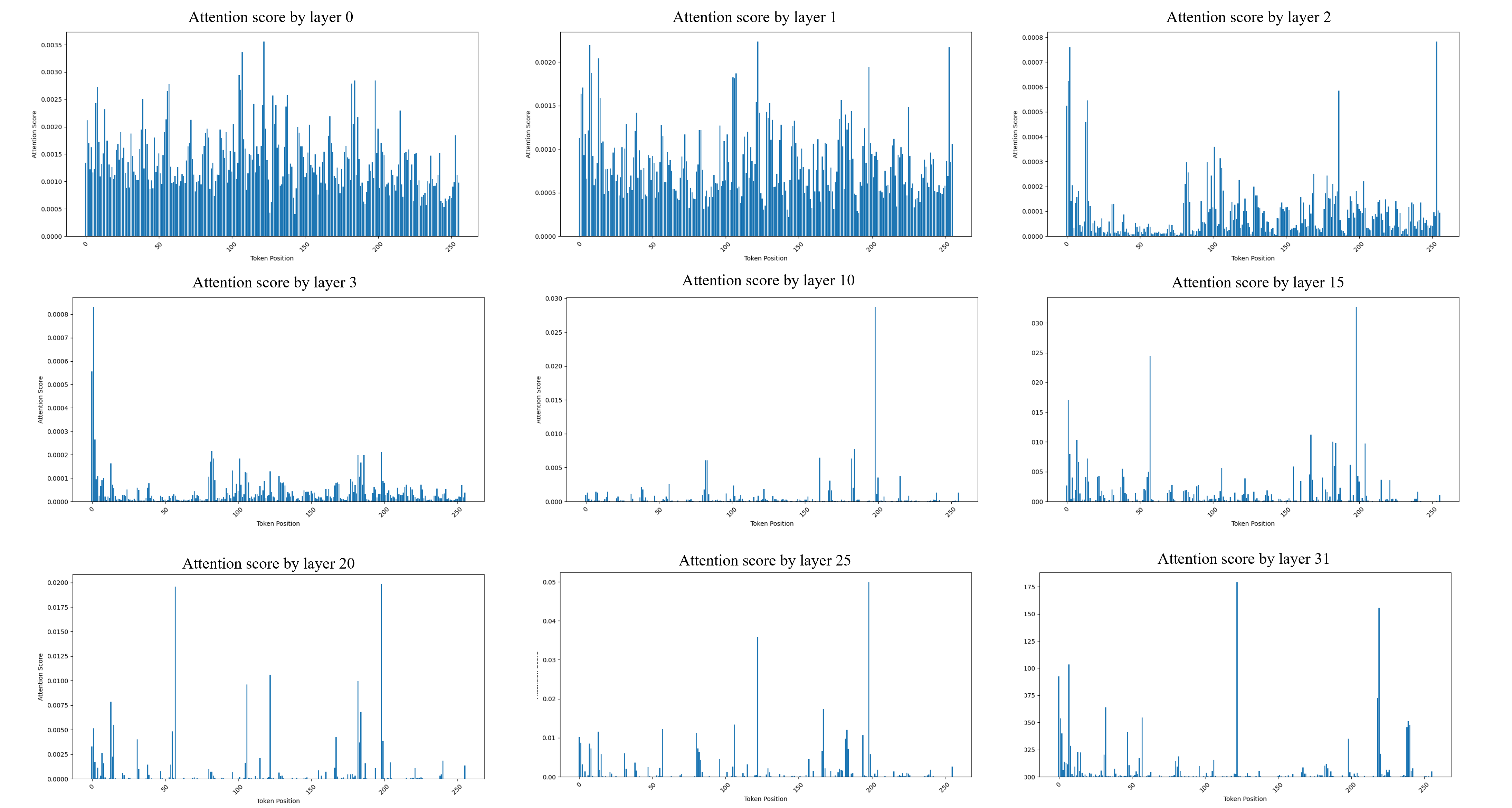}
    \caption{\textbf{Attention Score}: Attention score distributions across transformer layers in a VLA model. The scores are computed by averaging attention weights over heads and selecting the attention received by each token from the final query position. The results show increasing sparsity from the second layer onward, where attention becomes concentrated on a small subset of tokens.}
    \label{fig8_Attention_score}
\end{figure*}

To better understand the presence of visual redundancy in VLA models, we analyze the behavior of attention maps and attention scores across transformer layers. Figure~\ref{fig7_Attention_map} shows the average attention maps of VLA and VLM models across different layers. We observe that both types of models exhibit similar patterns: in early layers, attention is distributed relatively uniformly across visual tokens, while starting from the second layer, the attention maps become increasingly sparse and concentrated on fewer regions. This layer-wise transition suggests that redundancy accumulates early in the encoding process, making some tokens less informative in deeper layers.

To further quantify this sparsification effect, we examine the attention scores of visual tokens computed from the last transformer layer. Specifically, we extract the raw attention weights from the model outputs as follows:

{\small
\begin{quote}
\texttt{layer\_attention = layer\_outputs[1]} \\
\texttt{layer\_attention\_avg = torch.mean(layer\_attention, dim=1)[0]} \\
\texttt{attention\_score = layer\_attention\_avg[-1]}
\end{quote}
}

Here, the attention weights are averaged over all heads, and the final row corresponds to the attention received by each visual token when queried by the final position (e.g., the action token or decoder query). We use this vector as the attention score distribution.

As shown in Figure~\ref{fig8_Attention_score}, the attention scores are nearly uniform across token positions in the first two layers. However, starting from the second layer, we observe a clear increase in variance, with attention values increasingly concentrated on a small subset of tokens. This indicates a growing redundancy among visual tokens in deeper layers—a phenomenon also observed in recent studies on VLMs~\cite{chen2024image}. These observations provide empirical evidence for the existence of token-level redundancy in VLA models and motivate our token pruning strategy.





        




    

%
\section{Additional Experimental Details}
\label{sec:app_Additional_Experimental_Details}
\subsection{LIBERO Simulated Environment Benchmark}
LIBERO is a novel benchmark designed for studying knowledge transfer in multitask and lifelong robot learning. It addresses the challenge of benchmarking knowledge transfer capabilities in robot learning systems, with a focus on manipulation tasks that require both declarative knowledge (about objects and spatial relationships) and procedural knowledge (about motion and behaviors) .

LIBERO provides four main task suites, each designed to evaluate different aspects of knowledge transfer:

\paragraph{LIBERO-Spatial}
It contains 10 tasks that focus on transferring knowledge of spatial relationships. These tasks require robots to understand the spatial relationship between different objects and use this knowledge to complete the task. For example, the robot need to place objects according to a certain spatial layout, or navigate to the target position according to spatial clues in a complex environment. Through these tasks, its ability to master and apply the knowledge of spatial relationship is investigated.

\paragraph{LIBERO-Object}
It consists of 10 tasks that require transferring object-related knowledge. Robots are expected to identify different objects, comprehend their attributes (e.g., color, shape, material) and functions (e.g., tool usage, container functionality), and manipulate the objects accordingly. Examples include classifying objects based on their attributes or utilizing tools to perform specific tasks. These tasks serve to measure the robot's capability to transfer knowledge related to objects.

\paragraph{LIBERO-Goal}
It has 10 tasks that emphasize transferring goal-oriented knowledge. Robots must precisely comprehend the task objectives, determine the essential steps and strategies for achieving them. For example, it should be able to accurately prioritize goals in multi - task scenarios or break down complex goals into manageable sub - goals and accomplish them step by step. The evaluation aims to assess the robot's ability to transfer and apply goal - oriented knowledge effectively.

\paragraph{LIBERO-Long}
It has 10 tasks primarily designed to evaluate the robot's knowledge transfer ability over extended learning periods. These tasks typically involve learning and integrating knowledge across multiple tasks. The investigation focuses on whether the robot can effectively apply the experience, skills, and knowledge acquired from previous tasks to new subsequent tasks, thereby achieving continuous performance enhancement and improved adaptability. Furthermore, it emphasizes the accumulation, updating, transfer, and application of knowledge throughout the long - term learning process.

\begin{figure}[t]
    \centering
    \includegraphics[width=\linewidth]{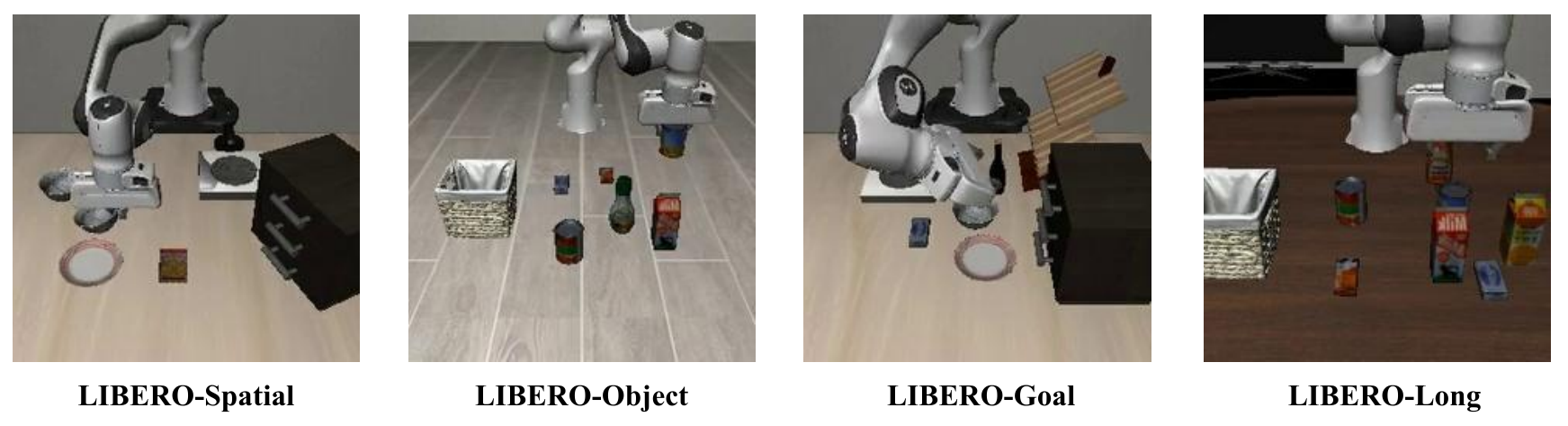}
    \caption{Sample Frame of Four Main Task Suites.}
    \label{fig9_Sampleframe}
\end{figure}



\section{Limitations and future works}
\label{sec:app_Limitations_and_future_works}
We propose \methodshort{}, the first training-free and plug-and-play acceleration framework that enables action reuse in VLA models. Although our approach maintains the performance of the model while greatly reducing the amount of modeling operations and the actuallatency, there are limitations and shortcomings in our approach. First of all we have only tested in a simulated environment, lacking further validation in the real world. Second, we only performe experimental validation of a single-arm robot. In the future, we will further validate the advantages of our approach in extending our method to more robotic arms with more degrees of freedom.

%% file: neurips_2025.bbl
\begin{thebibliography}{10}

\bibitem{ahn2022can}
M.~Ahn, A.~Brohan, N.~Brown, Y.~Chebotar, O.~Cortes, B.~David, C.~Finn, C.~Fu, K.~Gopalakrishnan, K.~Hausman, et~al.
\newblock Do as i can, not as i say: Grounding language in robotic affordances.
\newblock {\em arXiv preprint arXiv:2204.01691}, 2022.

\bibitem{bai2023qwen}
J.~Bai, S.~Bai, S.~Yang, S.~Wang, S.~Tan, P.~Wang, J.~Lin, C.~Zhou, and J.~Zhou.
\newblock Qwen-vl: A frontier large vision-language model with versatile abilities.
\newblock {\em arXiv preprint arXiv:2308.12966}, 1(2):3, 2023.

\bibitem{black2410pi0}
K.~Black, N.~Brown, D.~Driess, A.~Esmail, M.~Equi, C.~Finn, N.~Fusai, L.~Groom, K.~Hausman, B.~Ichter, et~al.
\newblock $\pi$0: A vision-language-action flow model for general robot control, 2024.
\newblock {\em URL https://arxiv. org/abs/2410.24164}.

\bibitem{brohan2022rt}
A.~Brohan, N.~Brown, J.~Carbajal, Y.~Chebotar, J.~Dabis, C.~Finn, K.~Gopalakrishnan, K.~Hausman, A.~Herzog, J.~Hsu, et~al.
\newblock Rt-1: Robotics transformer for real-world control at scale.
\newblock {\em arXiv preprint arXiv:2212.06817}, 2022.

\bibitem{cheang2024gr}
C.-L. Cheang, G.~Chen, Y.~Jing, T.~Kong, H.~Li, Y.~Li, Y.~Liu, H.~Wu, J.~Xu, Y.~Yang, et~al.
\newblock Gr-2: A generative video-language-action model with web-scale knowledge for robot manipulation.
\newblock {\em arXiv preprint arXiv:2410.06158}, 2024.

\bibitem{chen2024efficient}
J.~Chen, L.~Ye, J.~He, Z.-Y. Wang, D.~Khashabi, and A.~Yuille.
\newblock Efficient large multi-modal models via visual context compression.
\newblock In {\em The Thirty-eighth Annual Conference on Neural Information Processing Systems}, 2024.

\bibitem{chen2024image}
L.~Chen, H.~Zhao, T.~Liu, S.~Bai, J.~Lin, C.~Zhou, and B.~Chang.
\newblock An image is worth 1/2 tokens after layer 2: Plug-and-play inference acceleration for large vision-language models.
\newblock In {\em European Conference on Computer Vision}, pages 19--35. Springer, 2024.

\bibitem{chen2024fastv}
L.~Chen, H.~Zhao, T.~Liu, S.~Bai, J.~Lin, C.~Zhou, and B.~Chang.
\newblock An image is worth 1/2 tokens after layer 2: Plug-and-play inference acceleration for large vision-language models.
\newblock In {\em European Conference on Computer Vision}, pages 19--35. Springer, 2024.

\bibitem{chen2024ll3da}
S.~Chen, X.~Chen, C.~Zhang, M.~Li, G.~Yu, H.~Fei, H.~Zhu, J.~Fan, and T.~Chen.
\newblock Ll3da: Visual interactive instruction tuning for omni-3d understanding reasoning and planning.
\newblock In {\em Proceedings of the IEEE/CVF Conference on Computer Vision and Pattern Recognition}, pages 26428--26438, 2024.

\bibitem{chen2023executing}
X.~Chen, B.~Jiang, W.~Liu, Z.~Huang, B.~Fu, T.~Chen, and G.~Yu.
\newblock Executing your commands via motion diffusion in latent space.
\newblock In {\em Proceedings of the IEEE/CVF conference on computer vision and pattern recognition}, pages 18000--18010, 2023.

\bibitem{chi2023diffusion}
C.~Chi, Z.~Xu, S.~Feng, E.~Cousineau, Y.~Du, B.~Burchfiel, R.~Tedrake, and S.~Song.
\newblock Diffusion policy: Visuomotor policy learning via action diffusion.
\newblock {\em The International Journal of Robotics Research}, page 02783649241273668, 2023.

\bibitem{dao2022flashattention}
T.~Dao, D.~Fu, S.~Ermon, A.~Rudra, and C.~R{\'e}.
\newblock Flashattention: Fast and memory-efficient exact attention with io-awareness.
\newblock {\em Advances in neural information processing systems}, 35:16344--16359, 2022.

\bibitem{duan2024manipulate}
J.~Duan, W.~Yuan, W.~Pumacay, Y.~R. Wang, K.~Ehsani, D.~Fox, and R.~Krishna.
\newblock Manipulate-anything: Automating real-world robots using vision-language models.
\newblock {\em arXiv preprint arXiv:2406.18915}, 2024.

\bibitem{fang2023rh20t}
H.-S. Fang, H.~Fang, Z.~Tang, J.~Liu, C.~Wang, J.~Wang, H.~Zhu, and C.~Lu.
\newblock Rh20t: A comprehensive robotic dataset for learning diverse skills in one-shot.
\newblock {\em arXiv preprint arXiv:2307.00595}, 2023.

\bibitem{hou2024diffusion}
Z.~Hou, T.~Zhang, Y.~Xiong, H.~Pu, C.~Zhao, R.~Tong, Y.~Qiao, J.~Dai, and Y.~Chen.
\newblock Diffusion transformer policy.
\newblock {\em arXiv preprint arXiv:2410.15959}, 2024.

\bibitem{hu2022lora}
E.~J. Hu, Y.~Shen, P.~Wallis, Z.~Allen-Zhu, Y.~Li, S.~Wang, L.~Wang, W.~Chen, et~al.
\newblock Lora: Low-rank adaptation of large language models.
\newblock {\em ICLR}, 1(2):3, 2022.

\bibitem{jiang2023motiongpt}
B.~Jiang, X.~Chen, W.~Liu, J.~Yu, G.~Yu, and T.~Chen.
\newblock Motiongpt: Human motion as a foreign language.
\newblock {\em Advances in Neural Information Processing Systems}, 36:20067--20079, 2023.

\bibitem{khazatsky2024droid}
A.~Khazatsky, K.~Pertsch, S.~Nair, A.~Balakrishna, S.~Dasari, S.~Karamcheti, S.~Nasiriany, M.~K. Srirama, L.~Y. Chen, K.~Ellis, et~al.
\newblock Droid: A large-scale in-the-wild robot manipulation dataset.
\newblock {\em arXiv preprint arXiv:2403.12945}, 2024.

\bibitem{kim2025openvlaoft}
M.~J. Kim, C.~Finn, and P.~Liang.
\newblock Fine-tuning vision-language-action models: Optimizing speed and success.
\newblock {\em arXiv preprint arXiv:2502.19645}, 2025.

\bibitem{kim2024openvla}
M.~J. Kim, K.~Pertsch, S.~Karamcheti, T.~Xiao, A.~Balakrishna, S.~Nair, R.~Rafailov, E.~Foster, G.~Lam, P.~Sanketi, et~al.
\newblock Openvla: An open-source vision-language-action model.
\newblock {\em arXiv preprint arXiv:2406.09246}, 2024.

\bibitem{li2024inference}
K.~Y. Li, S.~Goyal, J.~D. Semedo, and J.~Z. Kolter.
\newblock Inference optimal vlms need only one visual token but larger models.
\newblock {\em arXiv preprint arXiv:2411.03312}, 2024.

\bibitem{li2024cogact}
Q.~Li, Y.~Liang, Z.~Wang, L.~Luo, X.~Chen, M.~Liao, F.~Wei, Y.~Deng, S.~Xu, Y.~Zhang, et~al.
\newblock Cogact: A foundational vision-language-action model for synergizing cognition and action in robotic manipulation.
\newblock {\em arXiv preprint arXiv:2411.19650}, 2024.

\bibitem{li2024evaluating}
X.~Li, K.~Hsu, J.~Gu, K.~Pertsch, O.~Mees, H.~R. Walke, C.~Fu, I.~Lunawat, I.~Sieh, S.~Kirmani, et~al.
\newblock Evaluating real-world robot manipulation policies in simulation.
\newblock {\em arXiv preprint arXiv:2405.05941}, 2024.

\bibitem{li2023vision}
X.~Li, M.~Liu, H.~Zhang, C.~Yu, J.~Xu, H.~Wu, C.~Cheang, Y.~Jing, W.~Zhang, H.~Liu, et~al.
\newblock Vision-language foundation models as effective robot imitators.
\newblock {\em arXiv preprint arXiv:2311.01378}, 2023.

\bibitem{liu2023libero}
B.~Liu, Y.~Zhu, C.~Gao, Y.~Feng, Q.~Liu, Y.~Zhu, and P.~Stone.
\newblock Libero: Benchmarking knowledge transfer for lifelong robot learning.
\newblock {\em Advances in Neural Information Processing Systems}, 36:44776--44791, 2023.

\bibitem{liu2024robomamba}
J.~Liu, M.~Liu, Z.~Wang, L.~Lee, K.~Zhou, P.~An, S.~Yang, R.~Zhang, Y.~Guo, and S.~Zhang.
\newblock Robomamba: Multimodal state space model for efficient robot reasoning and manipulation.
\newblock {\em arXiv preprint arXiv:2406.04339}, 2024.

\bibitem{liu2024rdt}
S.~Liu, L.~Wu, B.~Li, H.~Tan, H.~Chen, Z.~Wang, K.~Xu, H.~Su, and J.~Zhu.
\newblock Rdt-1b: a diffusion foundation model for bimanual manipulation.
\newblock {\em arXiv preprint arXiv:2410.07864}, 2024.

\bibitem{liu2024multi}
T.~Liu, L.~Shi, R.~Hong, Y.~Hu, Q.~Yin, and L.~Zhang.
\newblock Multi-stage vision token dropping: Towards efficient multimodal large language model.
\newblock {\em arXiv preprint arXiv:2411.10803}, 2024.

\bibitem{liu2024bidirectional}
Y.~Liu, J.~I. Hamid, A.~Xie, Y.~Lee, M.~Du, and C.~Finn.
\newblock Bidirectional decoding: Improving action chunking via closed-loop resampling.
\newblock {\em arXiv preprint arXiv:2408.17355}, 2024.

\bibitem{nair2022r3m}
S.~Nair, A.~Rajeswaran, V.~Kumar, C.~Finn, and A.~Gupta.
\newblock R3m: A universal visual representation for robot manipulation.
\newblock {\em arXiv preprint arXiv:2203.12601}, 2022.

\bibitem{oquab2023dinov2}
M.~Oquab, T.~Darcet, T.~Moutakanni, H.~V. Vo, M.~Szafraniec, V.~Khalidov, P.~Fernandez, D.~Haziza, F.~Massa, A.~El-Nouby, R.~Howes, P.-Y. Huang, H.~Xu, V.~Sharma, S.-W. Li, W.~Galuba, M.~Rabbat, M.~Assran, N.~Ballas, G.~Synnaeve, I.~Misra, H.~Jegou, J.~Mairal, P.~Labatut, A.~Joulin, and P.~Bojanowski.
\newblock Dinov2: Learning robust visual features without supervision, 2023.

\bibitem{o2024open}
A.~O’Neill, A.~Rehman, A.~Maddukuri, A.~Gupta, A.~Padalkar, A.~Lee, A.~Pooley, A.~Gupta, A.~Mandlekar, A.~Jain, et~al.
\newblock Open x-embodiment: Robotic learning datasets and rt-x models: Open x-embodiment collaboration 0.
\newblock In {\em 2024 IEEE International Conference on Robotics and Automation (ICRA)}, pages 6892--6903. IEEE, 2024.

\bibitem{park2024quantization}
S.~Park, H.~Kim, W.~Jeon, J.~Yang, B.~Jeon, Y.~Oh, and J.~Choi.
\newblock Quantization-aware imitation-learning for resource-efficient robotic control.
\newblock {\em arXiv preprint arXiv:2412.01034}, 2024.

\bibitem{pertsch2025fast}
K.~Pertsch, K.~Stachowicz, B.~Ichter, D.~Driess, S.~Nair, Q.~Vuong, O.~Mees, C.~Finn, and S.~Levine.
\newblock Fast: Efficient action tokenization for vision-language-action models.
\newblock {\em arXiv preprint arXiv:2501.09747}, 2025.

\bibitem{singh2023progprompt}
I.~Singh, V.~Blukis, A.~Mousavian, A.~Goyal, D.~Xu, J.~Tremblay, D.~Fox, J.~Thomason, and A.~Garg.
\newblock Progprompt: Generating situated robot task plans using large language models.
\newblock In {\em 2023 IEEE International Conference on Robotics and Automation (ICRA)}, pages 11523--11530. IEEE, 2023.

\bibitem{song2025pdvla}
W.~Song, J.~Chen, P.~Ding, H.~Zhao, W.~Zhao, Z.~Zhong, Z.~Ge, J.~Ma, and H.~Li.
\newblock Accelerating vision-language-action model integrated with action chunking via parallel decoding.
\newblock {\em arXiv preprint arXiv:2503.02310}, 2025.

\bibitem{touvron2023llama}
H.~Touvron, T.~Lavril, G.~Izacard, X.~Martinet, M.-A. Lachaux, T.~Lacroix, B.~Rozi{\`e}re, N.~Goyal, E.~Hambro, F.~Azhar, et~al.
\newblock Llama: Open and efficient foundation language models.
\newblock {\em arXiv preprint arXiv:2302.13971}, 2023.

\bibitem{wen2024diffusion}
J.~Wen, M.~Zhu, Y.~Zhu, Z.~Tang, J.~Li, Z.~Zhou, C.~Li, X.~Liu, Y.~Peng, C.~Shen, et~al.
\newblock Diffusion-vla: Scaling robot foundation models via unified diffusion and autoregression.
\newblock {\em arXiv preprint arXiv:2412.03293}, 2024.

\bibitem{wen2025tinyvla}
J.~Wen, Y.~Zhu, J.~Li, M.~Zhu, Z.~Tang, K.~Wu, Z.~Xu, N.~Liu, R.~Cheng, C.~Shen, et~al.
\newblock Tinyvla: Towards fast, data-efficient vision-language-action models for robotic manipulation.
\newblock {\em IEEE Robotics and Automation Letters}, 2025.

\bibitem{xu2025vlacache}
S.~Xu, Y.~Wang, C.~Xia, D.~Zhu, T.~Huang, and C.~Xu.
\newblock Vla-cache: Towards efficient vision-language-action model via adaptive token caching in robotic manipulation.
\newblock {\em arXiv preprint arXiv:2502.02175}, 2025.

\bibitem{yan2024dnact}
G.~Yan, Y.-H. Wu, and X.~Wang.
\newblock Dnact: Diffusion guided multi-task 3d policy learning.
\newblock {\em arXiv preprint arXiv:2403.04115}, 2024.

\bibitem{yang2024visionzip}
S.~Yang, Y.~Chen, Z.~Tian, C.~Wang, J.~Li, B.~Yu, and J.~Jia.
\newblock Visionzip: Longer is better but not necessary in vision language models.
\newblock {\em arXiv preprint arXiv:2412.04467}, 2024.

\bibitem{yue2024deer}
Y.~Yue, Y.~Wang, B.~Kang, Y.~Han, S.~Wang, S.~Song, J.~Feng, and G.~Huang.
\newblock Deer-vla: Dynamic inference of multimodal large language models for efficient robot execution.
\newblock {\em Advances in Neural Information Processing Systems}, 37:56619--56643, 2024.

\bibitem{zhai2023sigmoid}
X.~Zhai, B.~Mustafa, A.~Kolesnikov, and L.~Beyer.
\newblock Sigmoid loss for language image pre-training.
\newblock In {\em Proceedings of the IEEE/CVF international conference on computer vision}, pages 11975--11986, 2023.

\bibitem{zhang2025llava}
S.~Zhang, Q.~Fang, Z.~Yang, and Y.~Feng.
\newblock Llava-mini: Efficient image and video large multimodal models with one vision token.
\newblock {\em arXiv preprint arXiv:2501.03895}, 2025.

\bibitem{zhang2024vlabench}
S.~Zhang, Z.~Xu, P.~Liu, X.~Yu, Y.~Li, Q.~Gao, Z.~Fei, Z.~Yin, Z.~Wu, Y.-G. Jiang, and X.~Qiu.
\newblock Vlabench: A large-scale benchmark for language-conditioned robotics manipulation with long-horizon reasoning tasks, 2024.

\bibitem{zhang2024sparsevlm}
Y.~Zhang, C.-K. Fan, J.~Ma, W.~Zheng, T.~Huang, K.~Cheng, D.~Gudovskiy, T.~Okuno, Y.~Nakata, K.~Keutzer, et~al.
\newblock Sparsevlm: Visual token sparsification for efficient vision-language model inference.
\newblock {\em arXiv preprint arXiv:2410.04417}, 2024.

\bibitem{zitkovich2023rt}
B.~Zitkovich, T.~Yu, S.~Xu, P.~Xu, T.~Xiao, F.~Xia, J.~Wu, P.~Wohlhart, S.~Welker, A.~Wahid, et~al.
\newblock Rt-2: Vision-language-action models transfer web knowledge to robotic control.
\newblock In {\em Conference on Robot Learning}, pages 2165--2183. PMLR, 2023.

\end{thebibliography}
